%% file: main.tex
\pdfoutput=1
\documentclass[11pt]{article}

\usepackage{acl2023}

\usepackage{times}
\usepackage{latexsym}

\usepackage[T1]{fontenc}
\usepackage[utf8]{inputenc}

\usepackage{microtype}

\usepackage{inconsolata}

\usepackage{todonotes}
\usepackage{amsmath}
\usepackage{booktabs}
\usepackage{multicol}
\usepackage{graphicx}

\title{From \textit{chocolate bunny} to \textit{chocolate crocodile}:\\Do Language Models Understand Noun Compounds?}

\author{Jordan Coil$^1$ and Vered Shwartz$^{1,2}$ \\
$^1$ University of British Columbia~~~$^2$ Vector Institute for AI\\
{\tt jcoil93@students.cs.ubc.ca, vshwartz@cs.ubc.ca}}

\begin{document}
\maketitle
\begin{abstract}
\input{sections/0-abstract.tex}
\end{abstract}

\section{Introduction}
\label{sec:intro}
\input{sections/1-intro.tex}

\section{Background}
\label{sec:related_work}
\input{sections/5-related_work.tex}

\section{Noun Compound Interpretation}
\label{sec:nci}
\input{sections/2-nci.tex}

\section{Noun Compound Conceptualization}
\label{sec:ncc}
\input{sections/3-ncc.tex}

\section{Does GPT-3 Parrot its Training Data?}
\label{sec:ngrams}
\input{sections/4-ngrams.tex}

\section{Conclusion}
\label{sec:conclusion}
\input{sections/6-conclusion.tex}

\section{Limitations}
\label{sec:limitations}
\input{sections/7-limitations.tex}

\section{Ethical Considerations}
\label{sec:ethics}
\input{sections/8-ethics.tex}

% Uncomment for camera-ready version
\section*{Acknowledgements}

This work was funded, in part, by an NSERC USRA award, the Vector Institute for AI, Canada CIFAR AI Chairs program, an NSERC discovery grant, and a research gift from AI2. 

\bibliography{anthology,custom}
\bibliographystyle{acl_natbib}

\end{document}

%% file: sections/0-abstract.tex
Noun compound interpretation is the task of expressing a noun compound (e.g. \textit{chocolate bunny}) in a free-text paraphrase that makes the relationship between the constituent nouns explicit (e.g. \textit{bunny-shaped chocolate}). We propose modifications to the data and evaluation setup of the standard task \cite{hendrickx-etal-2013-semeval}, and show that GPT-3 solves it almost perfectly. We then investigate the task of noun compound conceptualization, i.e. paraphrasing a novel or rare noun compound. E.g., \textit{chocolate crocodile} is a crocodile-shaped chocolate. This task requires creativity, commonsense, and the ability to generalize knowledge about similar concepts. While GPT-3's performance is not perfect, it is better than that of humans---likely thanks to its access to vast amounts of knowledge, and because conceptual processing is effortful for people \cite{connell2012flexible}. Finally, we estimate the extent to which GPT-3 is reasoning about the world vs. parroting its training data. We find that the outputs from GPT-3 often have significant overlap with a large web corpus, but that the parroting strategy is less beneficial for novel noun compounds.

%% file: sections/1-intro.tex
\begin{figure}
    \centering
    \includegraphics[width=0.45\textwidth]{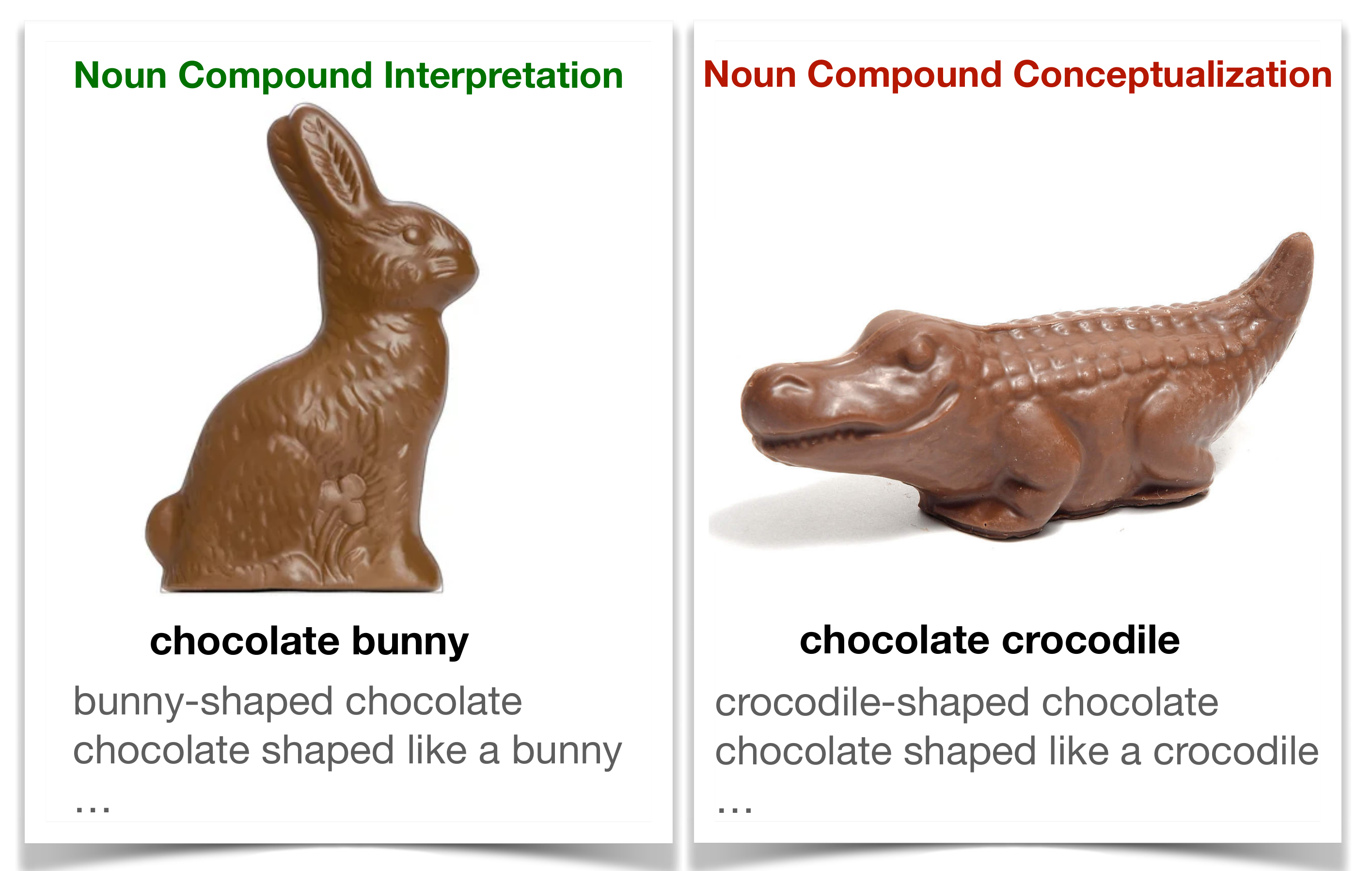}
    \caption{An example NC (input) and paraphrases (output) for each of the NCI and NCC tasks.}
    \label{fig:intro}
\end{figure}

Noun compounds (NCs) are prevalent in English, but most individual NCs are infrequent \cite{kim2007interpreting}. Yet, it is possible to derive the meaning of most NCs from the meanings of their constituent nouns. The task of noun compound interpretation (NCI) addresses this by explicitly uncovering the implicit semantic relation between the constituent nouns. We focus on the paraphrasing variant \cite{nakov2006using}, where the goal is to generate multiple paraphrases that explicitly express the semantic relation between the constituents. For example (Figure~\ref{fig:intro}), a \textit{chocolate bunny} is a ``chocolate shaped like a bunny''. 

Earlier methods for NCI represented NCs as a function their constituents' representations \cite[e.g.][]{van-de-cruys-etal-2013-melodi-supervised,shwartz-dagan-2018-paraphrase}. In recent years, pre-trained language models (PLMs) caused a paradigm shift in NLP. Such models are based on the transformer architecture \cite{vaswani2017attention}, which by design computes a word representation as a function of the representation of its context. Further, PLMs are pre-trained on vast amounts of text, which equips them with broad semantic knowledge \cite{rogers-etal-2020-primer}. Such knowledge may facilitate interpreting unseen NCs based on observed NCs that are semantically similar. Indeed, \newcite{ponkiya-etal-2020-looking} showed that a masked language model is useful for this task, and \newcite{shwartz-2021-long} demonstrated the utility of generative language models on this task. 

We formalize the experiments presented in \newcite{shwartz-2021-long} and evaluate generative models on NCI. We manually analyze and correct many problems with the standard SemEval 2013 task 4 dataset \cite{hendrickx-etal-2013-semeval}, and release a cleaned version of the dataset. Following the criticism in \newcite{shwartz-dagan-2018-paraphrase} on the task's dedicated evaluation metrics, we propose a more complete set of evaluation metrics including both automatic metrics and human evaluation. 

Our experiments show that a few-shot model based on GPT-3 \cite{gpt3} achieves near-perfect performance on the NCI test set. The impressive performance may be due to a combination of factors. First, it tends to memorize texts seen during pre-training \cite{carlini2022quantifying}, likely including partial or complete definitions of common NCs. Second, it has learned vast commonsense and world knowledge from its pre-training corpus, which---together with its ability to generalize---may be useful for interpreting less frequent NCs. 

To test the extent that GPT-3 reasons about its knowledge as opposed to memorizes definitions, we propose a second task: noun compound conceptualization (NCC). The setup is identical to NCI, but the NCs are rare or novel (e.g., \textit{chocolate crocodile} in Fig.~\ref{fig:intro}), requiring a model to come up with a plausible interpretation based on its existing knowledge. We construct a test set for this task based on data from \newcite{dhar-van-der-plas-2019-learning}. The results show that GPT-3 outperforms humans on NCC, presumably thanks to its fast access to a huge ``knowledge base'', and compared to the relative human slowness on this task \cite{connell2012flexible}.

Yet, compared to its performance on NCI, GPT-3's performance on NCC shows a significant drop. We thus quantify the extent that GPT-3 copies from its pre-training corpus when generating paraphrases for either NCI or NCC. We find that the generated paraphrases have significant overlap with a large web-based corpus, but that as expected, the copying strategy is less beneficial for NCC than for NCI. 

We anticipate that the cleaned dataset and proposed evaluation setup will be adopted by the research community for NCI, and hope to see further research on NCC.\footnote{The code and data are available at: \url{https://github.com/jordancoil/noun-compound-interpretation}}

%% file: sections/5-related_work.tex
\subsection{Noun Compound Interpretation}
\label{sec:related_work:nci}

Traditionally, NCI has been framed as a classification task into predefined relation labels. Datasets differed by the number of relations and their specificity level; from 8 prepositional relations \cite[e.g. \texttt{of}, \texttt{from}, etc.;][]{lauer1995}, to finer-grained inventories with dozens of relations \cite[e.g. \texttt{contains}, \texttt{purpose}, \texttt{time of};][]{kim-baldwin-2005-automatic, tratz-hovy-2010-taxonomy}. The classification approach is limited because even the larger relation inventories don't cover all possible relationships between nouns. In addition, each NC is classified to a single relation, although several relations may be appropriate.  E.g., \textit{business zone} is both a zone that \texttt{contains} businesses and a zone whose \texttt{purpose} is business \cite{shwartz-dagan-2018-paraphrase}. 

For these reasons, in this paper we focused on the task of interpreting noun compounds by producing multiple free-text paraphrases \cite{nakov2006using}. The reference paraphrases could be any text, but in practice they typically follow a ``[n$_2$] ... [n$_1$]'' pattern, where n$_1$ and n$_2$ are the constituent nouns. The main dataset for this task comes from SemEval 2013 task 4 \cite{hendrickx-etal-2013-semeval}, following a similar earlier task \cite{butnariu-etal-2009-semeval}. 

Earlier methods for this task reduced the paraphrasing task into a classification task to one of multiple paraphrase templates extracted from a corpus \cite{kim-nakov-2011-large,pasca-2015-interpreting,shwartz-dagan-2018-paraphrase}. \newcite{shwartz-dagan-2018-paraphrase} jointly learned to complete any item in the ([n$_1$], [n$_2$], paraphrase template) tuple, which allowed the model to generalize, predicting paraphrases for rare NCs based on similarity to other NCs. 

 More recently, \newcite{ponkiya-etal-2020-looking} showed that PLMs already capture this type of knowledge from their pre-training. They used an off-the-shelf T5 model to predict the mask substitutes in templates such as ``[n$_2$] [MASK] [n$_1$]'', achieving a small improvement over \newcite{shwartz-dagan-2018-paraphrase}. \newcite{shwartz-2021-long} further showed that supervised seq2seq models based on PLMs and a few-shot model based on GPT-3 yielded correct paraphrases for both common and rare NCs. 
 
\subsection{Forming and Interpreting new Concepts}
\label{sec:related_work:ncc}

Research in cognitive science studied how people interpret new noun-noun combinations such as \textit{cactus fish} \cite[e.g.][]{wisniewski,costello_keane,connell2012flexible}. While such combinations invite various interpretations,  there is usually a single preferred interpretation which is more intuitively understood. For example, a \textit{cactus fish} would more likely mean ``a fish that is spiky like a cactus'' than ``a fish that is green like a cactus'', because ``spiky'' is more characteristic of cacti than ``green'' \cite{costello_keane}.  

\newcite{connell2012flexible} constructed a set of 27 novel NCs and asked people to (1) judge the sensibility of an NC; and (2) come up with a plausible interpretation. The short response times for the sensibility judgment task indicated that participants relied on shallow linguistic cues as shortcuts, such as the topical relatedness between the constituent nouns. Response times in the interpretation generation task were longer, indicating that participants employed a slower process of mental simulation. Interpreting a new concept required building a detailed representation by re-experiencing or imagining the perceptual properties of the constituent nouns.

Computational work on plausibility judgement for NCs involves rare NCs \cite{lapata-lascarides-2003-detecting} and novel NCs \cite{dhar-van-der-plas-2019-learning}. The latter built a large-scale dataset of novel NCs by extracting positive examples from different decades in the Google Ngram corpus for training and testing. Negative examples were constructed by randomly replacing one of the constituents in the NC with another noun from the data. They proposed an LSTM-based model that estimates the plausibility of a target NC based on the pairwise similarity between the constituents of the target NC and other, existing NCs. For example, the candidate NC \textit{glass canoe} was predicted as plausible thanks to its similarity to \textit{glass boat}. 

In this paper, we go beyond plausibility judgement to the more complicated task of interpretation. In concurrent work, \newcite{li-etal-2022-systematicity} conducted similar experiments evaluating GPT-3's ability to define common and new noun compounds, as well as combinations of nonce words. They found no evidence that GPT-3 employs human-like linguistic principles when interpreting new noun compounds, and suggested it might be memorizing lexical knowledge instead. We further try to quantify the latter possibility in this work. 

Similarly to novel NCs, \newcite{pinter-etal-2020-will} look at novel blends from the NYTWIT corpus, collected automatically from a Twitter bot that tweets words published for the first time in the NYT \cite{pinter-etal-2020-nytwit}. For example, \textit{thrupple} is a blend of three and couple, used to describe ``A group of three people acting as a couple''. They found that PLMs struggled to separate blends into their counterparts. 

In a related line of work on creativity, researchers proposed models that coin new words from existing ones. \newcite{deri-knight-2015-make} generated new blends such as \textit{frenemy} (friend + enemy). \newcite{mizrahi-etal-2020-coming} generated new Hebrew words with an algorithm that is inspired by the human process of combining roots and patterns. 

%% file: sections/2-nci.tex
We first evaluate PLMs' ability to interpret existing noun compounds. We focus on the free-text paraphrasing version of NCI, as exemplified in Table~\ref{tab:orig_output_examples}. We use the standard dataset from SemEval 2013 Task 4 \cite{hendrickx-etal-2013-semeval}. We identified several problems in the dataset that we address in Sec~\ref{sec:nci_data}. We then trained PLM-based models on the revised dataset (Sec~\ref{sec:nci_methods}), and evaluated them both automatically and manually (Sec~\ref{sec:nci_results}).

\subsection{Data}
\label{sec:nci_data}
\input{sections/2-nci_data.tex}

\subsection{Methods}
\label{sec:nci_methods}
\input{sections/2-nci_methods.tex}

\subsection{Evaluation}
\label{sec:nci_results}
\input{sections/2-nci_results.tex}

%% file: sections/2-nci_data.tex
We manually reviewed the SemEval-2013 dataset and identified several major issues with the data quality. We propose a revised version of the dataset, with the following modifications.

\paragraph{Train-Test Overlap.} We discovered 32 NCs that appeared in both the training and test sets, and removed them from the test set. 

\paragraph{Incorrect Paraphrases.} We manually corrected paraphrases with superficial problems such as spelling or grammatical errors, redundant spaces, and superfluous punctuation. We also identified and removed NCs that were semantically incorrect. For example, \textit{rubber glove} was paraphrased to ``gloves has been made to get away from rubber'', perhaps due to the annotator mistaking the word \textit{rubber} for \textit{robber}. Finally, we found and removed a few paraphrases that contained superfluous or subjective additions, deviating from the instructions by \newcite{hendrickx-etal-2013-semeval}. For example, \textit{tax reduction} was paraphrased as ``reduction of tax \textit{hurts the economy}'', and \textit{engineering work} as ``work done \textit{by men} in the field of engineering''. Further, we discarded a total of 14 NCs from the training set and 11 NCs from the test set that had no correct paraphrases. In total, we removed 1,960 paraphrases from the training set and 5,066 paraphrases from the test set. 

\paragraph{``Catch-All'' Paraphrases.} The paraphrases in \newcite{hendrickx-etal-2013-semeval} were collected from crowdsourcing workers. An issue with the crowdsourcing incentive structure, is that it indirectly encourages annotators to submit any response, even when they are uncertain about the interpretation of a given NC. In the context of this dataset, this incentive leads to what we call ``catch-all'' paraphrases. Such paraphrases include generic prepositional paraphrases such as ``[n$_2$] of [n$_1$]'' (e.g. ``drawing of chalk''). For verbal paraphrases, the include generic verbs, such as ``[n$_2$] based on [n$_1$]'', ``[n$_2$] involving [n$_1$]'', ``[n$_2$] associated with [n$_1$]'', ``[n$_2$] concerned with [n$_1$]'', and ``[n$_2$] coming from [n$_1$]''. While these paraphrases are not always incorrect, they are also not very informative of the relationship between the constituent nouns. We therefor removed such paraphrases.\footnote{Another factor for the quality of paraphrases is the workers' English proficiency level. Writing non-trivial paraphrases requires high proficiency, and in 2013, it wasn't possible to filter workers based on native language on Mechanical Turk.}

\input{figures/data_stats.tex}

\paragraph{Data Augmentation.} To increase the size of the dataset in terms of paraphrases and facilitate easier training of models, we performed semi-automatic data augmentation. Using WordNet \cite{fellbaum2010wordnet}, we extended the set of paraphrases for each NC by replacing verbs with their synonyms and manually judging the correctness of the resultant paraphrase. We also identified cases were two paraphrases could be merged into additional paraphrases. For example, \textit{steam train} contained the paraphrases ``train powered by steam'' and ``train that operates using steam'', for which we added ``train operated by steam'' and ``train that is powered using steam''. Overall, we added 3,145 paraphrases to the training set and 3,115 to the test set.

We followed the same train-test split as the original dataset, but dedicated 20\% of the test set to validation. Table~\ref{tab:data_stats} displays the statistics of the NCI datasets. 

%% file: figures/data_stats.tex
\begin{table}[t]
\centering
\small
\setlength{\tabcolsep}{4pt}
\begin{tabular}{lrrrrrr}
\toprule
& \multicolumn{3}{c}{\textbf{Original}} & \multicolumn{3}{c}{\textbf{Revised}} \\
\midrule
& \textbf{train} & \textbf{dev} & \textbf{test} & \textbf{train} & \textbf{dev} & \textbf{test} \\
\midrule
\#NCs & 174 & 0 & 181 & 160 & 28 & 110 \\
\#paraphrases & 4,256 & 0 & 8,190 & 5,441 & 1,469 & 4,820 \\
\bottomrule
\end{tabular}
\caption{Statistics of the original SemEval 2013 dataset \cite{hendrickx-etal-2013-semeval} vs. our revised version (henceforth: the NCI dataset).}
\label{tab:data_stats}
\end{table}

%% file: sections/2-nci_methods.tex
We evaluate the performance of two representative PLM-based models on our revised version of the SemEval-2013 dataset (henceforth: the NCI dataset): a supervised seq2seq T5 model \cite{2020t5} and a few-shot prompting GPT-3 model \cite{gpt3}. 

\paragraph{Supervised Model.} We trained the seq2seq model from the Transformers package \cite{wolf2019huggingface}, using T5-large. We split each instance in the dataset into multiple training examples, with the NC as input and a single paraphrase as output. We used the default learning rate ($5\times10^{-5}$), batch size (16), and optimizaer (Adafactor). We stopped the training after 4 epochs when the validation loss stopped improving. During inference, we used top-p decoding \cite{holtzman2020curious} with $p = 0.9$ and a temperature of 0.7, and generated as many paraphrases as the number of references for a given NC.

\paragraph{Few-shot Model.} We used the \texttt{text-davinci-002} GPT-3 model available through the OpenAI API. We randomly sampled 10 NCs, each with one of its paraphrases, from the training set, to build the following prompt:
\begin{quote}
\texttt{Q: what is the meaning of <NC>?}\\\texttt{A:<paraphrase>}
\end{quote}
This prompt was followed by the same question for the target NC, leaving the paraphrases to be completed by GPT-3. We used the default setup of top-p decoding with $p=1$ and a temperature of 1. 

%% file: sections/2-nci_results.tex
\input{figures/orig_output_examples.tex}
\input{figures/orig_eval.tex}

We decided to deviate from the original evaluation setup of the SemEval 2013 dataset, which was criticized in \newcite{shwartz-dagan-2018-paraphrase}. We describe the original evaluation setup, and our proposed setup including automatic and manual evaluation. 

\paragraph{Original Evaluation Setup.} The original SemEval task was formulated as a ranking task. The paraphrases of each NC were ranked according to the number of annotators who proposed them. \newcite{hendrickx-etal-2013-semeval} introduced two dedicated evaluation metrics, an `isomorphic' score that measured the recall, precision, and order of paraphrases predicted by the systems, and a `non-isomorphic' score that disregarded the order. Both metrics rewarded systems for predicting shorter prepositional paraphrases (e.g. ``[n$_2$] of [n$_1$]''), that were in the set of paraphrases for many NCs, and were often ranked high because many annotators proposed them. 
For example, for the NC \textit{access road}, the catch-all paraphrase ``road for access'' was ranked higher than the more informative ``road that provides access''. Indeed, as noted in \newcite{shwartz-dagan-2018-paraphrase}, a baseline predicting a fixed set of common, generic paraphrases already achieves moderately good non-isomorphic score. In general, we do not see the benefit of the ranking system, since some of the most informative paraphrases are unique and are less likely to have been proposed by many annotators. Instead, we propose to use standard evaluation metrics for generative tasks, as we describe below. 

\paragraph{Automatic Evaluation.} Table~\ref{tab:orig_eval} (columns 2-4) displays the performance of T5 and GPT-3 on the test set using the following standard evaluation metrics for text generation tasks: the lexical overlap metrics ROUGE-L \cite{lin-2004-rouge} and METEOR \cite{lavie-agarwal-2007-meteor}, and the semantic-similarity metric BERT-Score \cite{bert}. These metrics compare the system generated paraphrases with the reference paraphrases, further motivating our data augmentation in Sec~\ref{sec:nci_data} (e.g., \newcite{lin-2004-rouge} found that considering multiple references improves ROUGE's correlation with human judgements). For each metric $\operatorname{m}$, we compute the following score over the test set T:
\begin{align*}
s = \operatorname{mean}_{\text{nc} \in T} \bigr[ & \operatorname{mean}_{\text{p} \in \operatorname{system}(\text{nc})} \\
& \operatorname{max}_{\text{r} \in \operatorname{references}(\text{nc})} \operatorname{m}(p, r) \bigr]
\end{align*}

\noindent In other words, we generate a number of paraphrases equal to the number of reference paraphrases, then find the most similar reference for each of the generated paraphrases, and average across all paraphrases for each NC in the test set.

The automatic metrics show a clear preference to T5. However, upon a closer look at the outputs of each model, it seems that T5 generated paraphrases that more closely resembled the style and syntax of the references, as expected from a supervised model, but the paraphrases were not ``more correct'' than those outputted by GPT-3. For example, in Table~\ref{tab:orig_output_examples}, the paraphrase generated by GPT-3 for \textit{reflex action} is correct but doesn't follow the syntax of the references in the training data ([n$_2$] ... [n$_1$]). The T5-generated paraphrase follows that syntax but generates the generic and inaccurate paraphrase ``action performed to perform reflexes''. More broadly, lexical overlap based metrics such as ROUGE and METEOR penalize models for lexical variability.

\paragraph{Human Evaluation.} To assess the quality of predictions in a more reliable manner, we turn to human evaluation. We used Amazon Mechanical Turk (MTurk) and designed a human intelligence task (HIT) which involved displaying an NC along with 10 generated paraphrases, 5 from GPT-3 and 5 from T5, randomly shuffled. We asked workers to indicate for each paraphrase whether they deemed it acceptable or not. Each HIT was to be performed by 3 workers, and acceptability was measured using majority voting. To ensure the quality of workers, we required that workers reside in the US, Canada, or the UK, and that they had an acceptance rate of at least 99\% for all prior HITs. We also required them to pass a qualification task that resembled the HIT itself. We paid each worker \$0.10 per task, which yielded an approximate hourly wage \$15.

The last column in Table~\ref{tab:orig_eval} presents the results of the human evaluation in terms of percentage of paraphrases deemed acceptable by a majority of human evaluators. GPT-3 performed remarkably well with over 95\% of generated paraphrases deemed acceptable by a majority of human evaluators. In contrast to the automatic metrics, T5 fared much worse on human evaluation, and human annotators judged a third of T5 outputs as incorrect.

%% file: figures/orig_output_examples.tex
\begin{table*}[t]
\centering
\small
\begin{tabular}{lll}
\toprule
\textbf{NC} & \textbf{GPT-3} & \textbf{T5}\\
\midrule
\textit{access road} & road that provides access & road for access \\
\textit{reflex action} & a sudden, involuntary response to a stimulus & action performed to perform reflexes \\
\textit{sport page} & a page in a publication that is devoted to sports & page dedicated to sports \\
\textit{computer format} & the way in which a computer organizes data & format used in computers \\
\textit{grief process} & process of grieving or mourning & process that a grief sufferer experiences \\
\bottomrule
\end{tabular}
\caption{Example paraphrases generated using GPT-3 and T5 for NCs in the revised SemEval 2013 test set.}
\label{tab:orig_output_examples}
\end{table*}

%% file: figures/orig_eval.tex
\begin{table*}[t]
\centering
\small
\begin{tabular}{lrrrrrr}
\toprule
\textbf{Method} & \multicolumn{1}{c}{\textbf{METEOR}} & \multicolumn{1}{c}{\textbf{ROUGE-L}} & \multicolumn{1}{c}{\textbf{BERTScore}} &  \multicolumn{1}{c}{\textbf{Human}}\\
\midrule
T5 & \textbf{69.81} & \textbf{65.96} & \textbf{95.31}   & 65.35 \\
GPT-3 & 56.27 & 47.31 & 91.94 & \textbf{95.64} \\
\bottomrule
\end{tabular}
\caption{Performance of the T5 and GPT-3 models on the revised SemEval 2013 test set.}
\label{tab:orig_eval}
\end{table*}

%% file: sections/3-ncc.tex
GPT-3's impressive success at interpreting existing noun compounds is related to PLMs' ability to associate nouns with their hypernyms \cite{ettinger-2020-bert} and to generate accurate definitions for terms \cite{shwartz-etal-2020-unsupervised}. Such models are trained on vast amounts of texts, including said definitions, and the target NC itself occurring alongside contexts that indicate its meaning. Humans are different in their ability to interpret NCs. We can often rely on a single context, or no context at all, to have at least an educated guess at the meaning of a new NC. We are capable of representing new concepts by ``mentally manipulating old ones'' \cite{connell2012flexible}, e.g. coming up with a plausible interpretation for \textit{chocolate crocodile} based on similar concepts such as \textit{chocolate bunny}. Prior work on NCI simulated this by training a model to jointly predict a paraphrase as well as answer questions such as ``what can chocolate be shaped like?'' \cite{shwartz-dagan-2018-paraphrase}. We are interested in learning whether PLMs already do this implicitly, or more broadly, to what extent can PLMs interpret new noun compounds?

Inspired by studies in cognitive science about ``conceptual combination'' \cite{wisniewski,costello_keane}, we define the task of Noun Compound Conceptualization (NCC). NCC has the same setup as NCI (\S\ref{sec:nci}), but the inputs are rare or novel noun compounds. The task thus requires some level of creativity and the ability to make sense of the world. We first describe the creation of the NCC test set (Sec~\ref{sec:ncc_data}). We evaluate the best model from Sec~\ref{sec:nci_methods} on the new test set, and present the results in Sec~\ref{sec:ncc_results}. 

\subsection{Data}
\label{sec:ncc_data}
\input{sections/3-ncc_data.tex}

\begin{figure*}
    \centering
    \includegraphics[width=\textwidth]{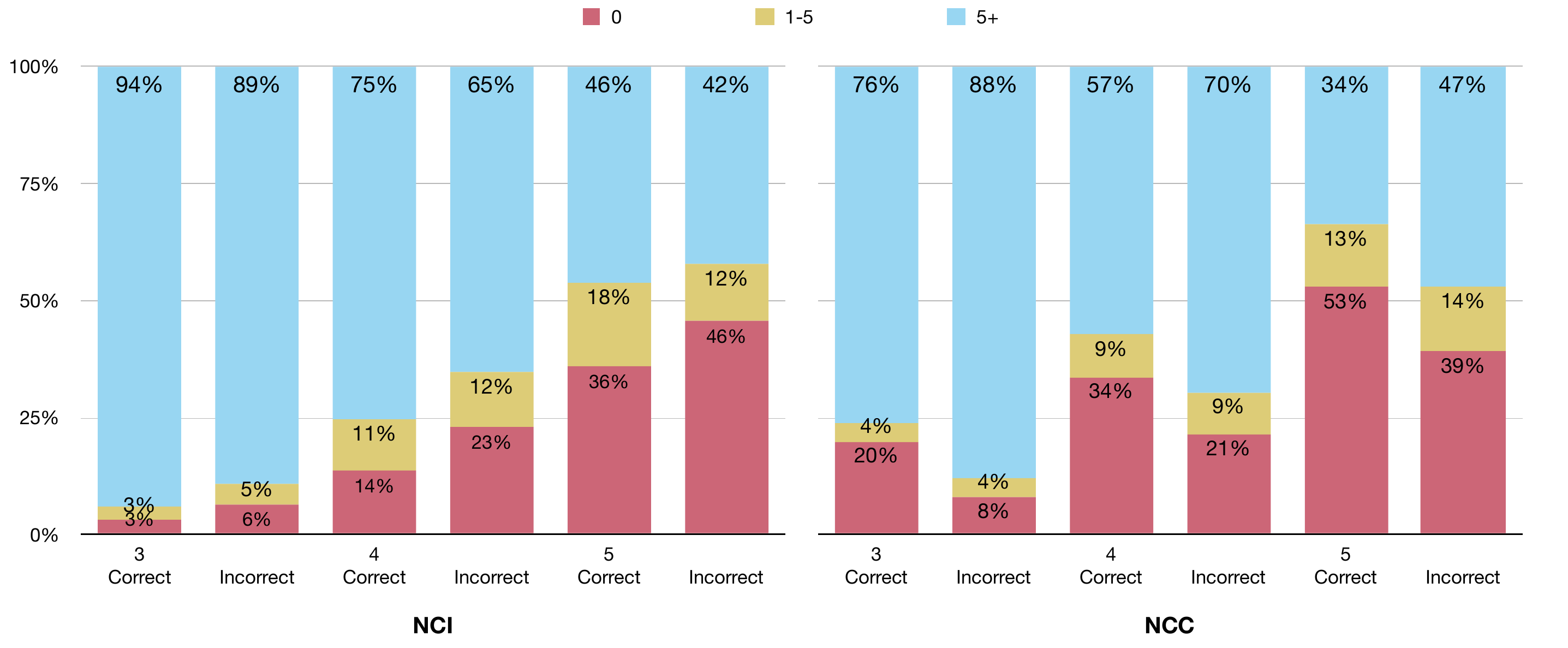}
    \caption{The percent of n-grams among the generated paraphrases (for $n = \{3, 4, 5\}$) that occur in the C4 corpus 0, 1-5, or 5+ times, for each of the NCI and NCC test sets, grouped by correct vs. incorrect generated paraphrases.}
    \label{fig:ngram_overlap}
\end{figure*}

\subsection{Evaluation}
\label{sec:ncc_results}
\input{sections/3-ncc_results.tex}

%% file: sections/3-ncc_data.tex
We construct a new test set consisting of novel or rare NCs. The guidelines for adding an NC for the test set are that: (a) humans could easily make sense of it; but (b) it is infrequent in or completely absent from the web. 

\paragraph{Noun Compounds.} The main source for the test set is a dataset from \newcite{dhar-van-der-plas-2019-learning}. They proposed the task of classifying an unseen sequence of two nouns to whether it can form a plausible NC or not. The data was created by extracting noun-noun bigrams from the Google Ngram corpus \cite{brants2006web}. To simulate novel NCs, the models were trained on bigrams that only appeared in the corpus until the year 2000 and evaluated on bigrams that only appeared after 2000. Since GPT-3 was trained on recent data, we had to make sure that we only include the most infrequent NCs. We thus further refined the data from \newcite{dhar-van-der-plas-2019-learning} by including only the 500 most infrequent NCs based on their frequency in a large-scale text corpus, C4 \cite{2020t5}. We then semi-automatically filtered out named entities, compounds that were part of larger expressions, and NCs with spelling errors. Finally, we manually chose only the NCs for which we could come up with a plausible interpretation, leaving us with 83 NCs in total. 

We added 22 more NCs that we extracted in a similar manner from the Twitter sentiment 140 dataset \cite{go2009twitter}. We expected to find more ``ad-hoc'' NCs in tweets than in more formal texts such as news. Due to the age and size of this dataset, we filtered the NCs based on frequency in C4, setting the threshold to 250 occurances. Overall, our NCC test set contains a total of 105 NCs.

\paragraph{Paraphrases.} We collected reference paraphrases for the NCC test set using MTurk. We showed workers the target NC and asked them to paraphrase the NC or give their best estimate if they are unfamiliar with the NC. We used the same qualifications as in Sec~\ref{sec:nci_results}, and paid \$0.12 per HIT.  

%% file: sections/3-ncc_results.tex
\input{figures/diff_eval.tex}

We focus on GPT-3 due to its almost perfect performance on NCI. We evaluated GPT-3 on the NCC test set using the few-shot setup described in Sec~\ref{sec:nci_methods}. We selected the few-shot examples from the NCI training set. 

We focus on human evaluation (as described in Sec~\ref{sec:nci_results}), which is more reliable than automatic metrics. We asked workers to judge the validity of both human-written and GPT-3 generated paraphrases. 

Table~\ref{tab:diff_eval} shows that GPT-3 performs significantly better than humans at this task. GPT-3 benefits from access to huge amounts of data. We conjecture that even though the target NCs are rare in its training data, it likely observed similar NCs, and is able to generalize and make sense of new concepts. At the same time, while humans are in general capable of coming up with a plausible interpretation for an unfamiliar concept, it is an effortful and cognitively taxing task. We hypothesize that in a setup other than crowdsourcing, i.e. given more time or incentive, human performance may increase.

Compared to its performance on NCI, GPT-3's performance on NCC shows a significant drop. This may suggest that GPT-3 struggles to reason about certain rare NCs, which we investigate in the next section.

%% file: figures/diff_eval.tex
\begin{table}[t]
\centering
\small
\begin{tabular}{lrr}
\toprule
\textbf{Test Set} & \multicolumn{1}{c}{NCI} & \multicolumn{1}{c}{NCC} \\
\midrule
Human Performance & \multicolumn{1}{c}{-} & 73.33 \\
GPT-3 & 95.64 & 83.81 \\
\bottomrule
\end{tabular}
\caption{Human evaluation performance (percent of correct paraphrases) of paraphrases proposed by people or generated by GPT-3 for the NCI and NCC test sets.}
\label{tab:diff_eval}
\end{table}

%% file: sections/4-ngrams.tex
While GPT-3 performs fairly well on NCC, looking at failure cases brings up interesting observations. For example, one of its responses for \textit{chocolate crocodile} was ``A large, aggressive freshwater reptile native to Africa''. This response seems to have ignored the \textit{chocolate} part of the NC entirely, and opted to provide an answer to ``What is a crocodile?''. Much like a student who doesn't know the answer to a question so instead regurgitates everything they memorized about the topic in hopes that it will include the correct answer.\footnote{A similar phenomenon was also demonstrated in concurrent work from \newcite{li-etal-2022-systematicity}. They showed that for instance, GPT-3 defines a \textit{banana table} as a \textit{banana} rather than a \textit{table}, differently from humans.}

To quantify the extent to which GPT-3 may be parroting its training corpus, we look at n-gram overlap between GPT-3's generated paraphrases and the large-scale web-based corpus C4 \cite{2020t5}.\footnote{We don't have access to the GPT-3 training corpus, but it included Common Crawl, web texts, books, and Wikipedia. C4 \cite{2020t5} is a colossal, cleaned version of Common Crawl, thus it is the closest to GPT-3's training corpus.}

Figure~\ref{fig:ngram_overlap} displays the percents of n-grams among the generated paraphrases (for $n = \{3, 4, 5\}$) that occur in the C4 corpus 0, 1-5, or 5+ times, for each of the NCI and NCC test sets. The results are presented separately for paraphrases deemed correct and incorrect by human evaluators.

We learn several things from Figure~\ref{fig:ngram_overlap}. First, the generated paraphrases often had significant overlap with the corpus (34-94\%). As expected, trigrams are copied more than 4-grams, which are copied more than 5-grams, as those tend to be rarer. 

Second, for the NCI test set, for each $n$, we see that n-grams from the correct paraphrases are copied from the web more often than n-grams from the incorrect paraphrases. The trend is reversed for NCC, where incorrect paraphrases are copied from the web more often than correct ones. Naturally, the copying strategy is less useful for NCC, which requires reasoning about new concepts. When GPT-3 generates correct paraphrases for NCC, their n-grams tend to not appear in the web at all. 

We reach a similar conclusion by looking at the percent of n-grams in correct vs. incorrect paraphrases that are copied from the web. The vast majority of n-grams copied from the web (97\%) for the NCI test set were correct, as opposed to only 80\% for NCC.

%% file: sections/6-conclusion.tex
We evaluated PLMs on their ability to paraphrase existing and novel noun compounds. For interpretation of existing NCs (NCI), we released a cleaned version of the SemEval 2013 dataset, with manual correction and automatic augmentation of paraphrases, and proposed additional evaluation metrics to overcome limitations described in prior work. GPT-3 achieved near perfect performance on this new test set. We then investigated the task of noun compound conceptualization (NCC). NCC evaluates the capacity of PLMs to interpret the meaning of new NCs. We showed that GPT-3 still performs reasonably well, but its success can largely be attributed to copying definitions or parts of definitions from its training corpus. 

%% file: sections/7-limitations.tex
\paragraph{Human performance on NCC.} The human accuracy on NCC was 73\%, compared to 83\% for GPT-3. We know from cognitive science research that humans are capable of forming new concepts based on existing ones \cite{connell2012flexible}. Moreover, we manually selected NCs in the NCC test set that we could come up with a plausible interpretation for. The fact that 27\% of the paraphrases proposed by MTurk workers were judged as incorrect could be explained by one of the following. The first explanation has to do with the limitations of crowdsourcing. To earn enough money, workers need to perform tasks quickly, and conceptualization is a slow cognitive process. On top of that, a worker that has already spent considerable amount of time trying to come up with a plausible interpretation for a new NC, is incentivized to submit any answer they managed to come up with, regardless of its quality. Skipping a HIT means lost wages. In a different setup, we hypothesize that human performance may increase for this task. 

The second explanation has to do with the evaluation setup. We asked people to judge paraphrases as correct or incorrect. Upon manual examination of a sample of the human-written paraphrases, we observed a non-negligible number of reasonable (but not optimal) paraphrases that were annotated as incorrect. For future work, we recommend doing a more nuanced human evaluation that will facilitate comparing the outputs of humans and models along various criteria. 

\paragraph{The work focuses only on English.} Our setup and data construction methods are fairly generic and we expect it to be straightforward to adapt them to other languages that use noun compounds. With that said, languages such as German, Norwegian, Swedish, Danish, and Dutch write noun compounds as a single word. Our methods will not work on these languages without an additional step of separating the NC into its constituent nouns, similar to unblending blends \cite{pinter-etal-2020-will}. In the future, we would like to investigate how well PLMs for other languages perform on NCI and NCC, especially for low-resource languages. 

\paragraph{Limitations of automatic metrics for generative tasks.} Automatic metrics based on n-gram overlap are known to have low correlation with human judgements on various NLP tasks \cite{novikova-etal-2017-need}. In particular, they penalize models for lexical variability. To mitigate this issue, we semi-automatically expanded the set of reference paraphrases using WordNet synonyms. Yet, we still saw inconsistencies with respect to the automatic metrics and human evaluation on NCI. The automatic metrics showed a clear preference to T5, which thanks to the supervision, learned to generate paraphrases that more closely resembled the style and syntax of the references. GPT-3's paraphrases, which were almost all judged as correct by human annotators, were penalized by the automatic metrics for their free form (e.g., they didn't always include the constituent nouns). For this reason, we focused only on human evaluation for NCC.

%% file: sections/8-ethics.tex
\paragraph{Data Sources.} All the datasets and corpora used in this work are publicly available. The cleaned version of the NCI dataset is based on the existing SemEval 2013 dataset \cite{hendrickx-etal-2013-semeval}. The NCs for the new NCC test set were taken from another publicly-available dataset \cite{dhar-van-der-plas-2019-learning}, based on frequencies in the Google Ngram corpus \cite{brants2006web}. To quantify Ngram overlap, we used the Allen AI version of the C4 corpus \cite{2020t5,dodge-etal-2021-documenting} made available by the HuggingFace Datasets package.\footnote{\url{https://huggingface.co/datasets/c4}}

\paragraph{Data Collection.} We performed human evaluation using Amazon Mechanical Turk. We made sure annotators were fairly compensated by computing an average hourly wage of \$15, which is well above the US minimum wage. We did not collect any personal information from annotators. 

\paragraph{Models.} The models presented in this paper are for a low-level NLP task rather than for an application with which users are expected to interact directly. The generative models are based on PLMs, which may generate offensive content if prompted with certain inputs.